\documentclass[11pt,letterpaper]{article}
\usepackage{cogsys}
\usepackage[T1]{fontenc}
\usepackage{times}
\usepackage[pdftex]{graphicx} 

\usepackage{natbib}
\cogsysheading{X}{20XX}{1-6}{X/20XX}{X/20XX}

\ShortHeadings{Online Learning of HTN Methods}
              {Y.\ Xu, H.\ Munoz-Avila}

\usepackage{balance} 

\usepackage{makecell}

\usepackage{mathtools}

\usepackage{amssymb}

\usepackage{algorithm}
\usepackage{algpseudocode}%

\usepackage{subcaption}

\usepackage{listings}
\begin{document}

\newcommand{\D}{$\mathcal{D}$}

\title{Online Learning of HTN Methods
for integrated LLM-HTN Planning}

\author{Yuesheng Xu}{yux220@lehigh.edu}
\author{H\'ector Mu\~noz-Avila}{hem4@lehigh.edu}
\address{Computer Science \& Engineering,
Lehigh University;
Bethlehem, PA  18015-3084  USA}
\vskip 0.2in

 \begin{abstract}
We present online learning of Hierarchical Task Network (HTN) methods in the context of integrated  
HTN planning and LLM-based chatbots. Methods indicate when and how to decompose tasks into subtasks.  
Our method learner is built on top of the ChatHTN planner. ChatHTN queries  
ChatGPT to generate a decomposition of a task into primitive tasks when no applicable method for the task is  
available. In this work, we extend ChatHTN. Namely, when ChatGPT generates a task decomposition,  
ChatHTN learns from it, akin to memoization. However, unlike memoization, it learns a generalized  
method that applies not only to the specific instance encountered, but to other instances of the same task..  
We conduct experiments on two domains and demonstrate that our online learning procedure reduces the number of calls to ChatGPT while solving at least as many problems, and in some cases, even more.
\end{abstract}

\section{Introduction}

Tasks in Hierarchical Task Networks (HTNs) represent activities to be performed,  
such as searching and rescuing survivors in an area affected by a natural disaster.  
HTN planners generate solution plans (i.e., sequences of actions) by  
recursively decomposing these complex tasks into simpler tasks. For instance,  
they may decompose the search-and-rescue task into subtasks that search for and rescue survivors  
in specific locations within the disaster area.  
A plan is formed as so-called primitive tasks are generated.  
Primitive tasks are atomic tasks achieved  
by an associated action. An example of such a primitive task is  
unloading a survivor from a drone into a safe haven.

HTN planning is a frequently studied topic because of its applications in real-world domains  
(e.g., \citep{nau2005applications}), including military planning \citep{Donaldso2014}, game  
AI \citep{smith1998success,Verweij2007}, and UAV control \citep{cardoso2017multi}.  
HTN's stratified representation of tasks of varied complexity has also been exploited in cognitive  
architectures \citep{laird2019soar,langley2006unified}. This is a natural match, given that  
stratified representations mimic how humans learn: starting with simpler tasks and  
progressively learning more sophisticated tasks \citep{choi2005learning}.

Despite these successes, a major stumbling block in the adoption of HTN planning  
is the need to supply a complete and correct set of methods for a wide range of tasks.  
A method indicates when and how to decompose a task.  
In recent work \citep{munozavila2025chathtninterleavingapproximatellm}, we reported on ChatHTN, an HTN planner  
that queries ChatGPT to generate  
a decomposition of the task into primitive tasks when no applicable method is available.  
Then, HTN planning proceeds using the newly generated method and queries ChatGPT again only when needed.  
ChatHTN is provably sound: any solution it generates is guaranteed to correctly solve the given problem.

In this work, we extend ChatHTN. Namely, when ChatGPT generates a task decomposition,  
ChatHTN learns from it, akin to memoization \citep{michie1968memo}.  
But unlike memoization, it learns a generalized method that applies not only to the specific instance encountered, but to many others.  
Furthermore, the methods are learned in such a way that ChatHTN's  
soundness is preserved.  
Our aim is to reduce the number of calls to ChatGPT. Reducing calls to ChatGPT (or any other LLM-based chatbot) is a  
highly desirable property due to both monetary costs and response time: a chatbot query takes several seconds, while method-based decomposition completes in a few hundred nanoseconds.\footnote{Setting up the domains for this work  
and running the experiments incurred costs exceeding US \$300. In our experiments, methods have around  
15 applicability conditions and subtasks. We ran these experiments on a Mac with an M1 processor.}

The following are the contributions of this work:

\begin{enumerate}
    \item We present the online HTN learning problem with chatbots.
    \item We introduce a method learning procedure that preserves the  soundness guarantees of ChatHTN and guarantee the correctness of the learned methods themselves. 
    \item We present our empirical evaluation  showing that that method learning reduces the calls to ChatGPT  while solving at least as many problems, and in some cases, even more. Our experiments span two benchmark domains from the HTN planning literature.
\end{enumerate}

The rest of the paper is structured as follows.  
First, we present a sample scenario illustrating the  
advantages of online learning of methods for ChatHTN.  
Next, we describe the ChatHTN planner.  
Afterwards, we present our procedure for learning methods and introduce the notion of  
termination methods. We then discuss the empirical evaluation.  
Finally, we review related work and  
make concluding remarks.

\section{Example Scenario}


To illustrate both ChatHTN and the need for online method learning, we present an example based on the search-and-rescue domain \citep{cottam1998knowledge}. In this domain, a catastrophe has occurred in a specific area. A safe haven is established with medical services, food, shelter, etc. The planning task is to search for and rescue survivors, who need to be found and relocated to the safe haven.

\begin{figure}[htb]
\centering
\begin{lstlisting}
searchANDrescue(Alpha)
    !scanArea(Zulu)
    checkSurvivors(Zulu)
        rescueSurvivor(Maria,Zulu)  <<<<
        checkSurvivors(Zulu)
    searchANDrescue(Alpha)
\end{lstlisting}
\caption{\emph{Initial decomposition in the search-and-rescue domain. ChatHTN does
not have a method decomposing the task rescueSurvivor(Maria,Zulu)} - annotated with "$<<<<$".}
\label{initialHierarchy}
\end{figure}

Figure \ref{initialHierarchy} shows the top-level task \emph{searchANDrescue(Alpha)}, where Alpha is the catastrophe area. This task is decomposed into three subtasks: \emph{!scanArea(Zulu), checkSurvivors(Zulu), searchANDrescue(area)}, where \emph{Zulu} is a location in the area, using the searchANDrescueM2 method shown in Table \ref{tab:htn-methods}. Details of these methods will be presented later. The exclamation mark in front of a task indicates a primitive task accomplished by operators. Thus, \emph{checkSurvivors(Zulu)} must be decomposed. This task is decomposed into the subtasks: \emph{rescueSurvivor(Maria,Zulu), checkSurvivors(Zulu)}.

Suppose there is no method applicable for the \emph{rescueSurvivor(Maria,Zulu)} task. To address this, ChatHTN employs a two-step prompt chaining strategy to query ChatGPT for a task decomposition, first with the task and current state, and then with the same information augmented by ChatGPT’s initial response; see~\cite{munozavila2025chathtninterleavingapproximatellm} for prompt details. Suppose that ChatGPT generates the decomposition for the \emph{rescueSurvivor(Maria,Zulu)} task shown in Figure \ref{refinedHierarchy}. ChatHTN then continues decomposing \emph{checkSurvivors(Zulu)} into \emph{rescueSurvivor(John,Zulu), checkSurvivors(Zulu)}. Here, ChatHTN must again prompt ChatGPT to decompose \emph{rescueSurvivor(John,Zulu)}.  

In contrast, our system learns from the decomposition generated for \emph{rescueSurvivor(Maria,Zulu)} and produces the method \emph{rescueSurvivorM2} shown in the table. It then applies this method to \emph{rescueSurvivor(John,Zulu)} to decompose it as shown in Figure \ref{refinedHierarchy}. Notably, this occurs despite the decomposition being learned from a different instance of the same task and in a different state.

\begin{figure}[htb]
\centering
\begin{lstlisting}
searchANDrescue(Alpha)
    !scanArea(Zulu)
    checkSurvivors(Zulu)
        rescueSurvivor(Maria,Zulu)
            !fly(Drone01,safeHaven,Zulu)
            !pickUpSurvivor(Drone01,Maria,Zulu)
            !fly(Drone01,Zulu,safeHaven)
            !dropSurvivor(Drone01,Maria,SafeHeaven)
        checkSurvivors(Zulu)
            rescueSurvivor(John,Zulu)    <<<<
                !fly(Drone01,safeHaven,Zulu)
                !pickUpSurvivor(Drone01,John,Zulu)
                !fly(Drone01,Zulu,safeHaven)
                !dropSurvivor(Drone01,John,SafeHeaven)
            checkSurvivors(Zulu)
                !doNothing()
    searchANDrescue(Alpha)
        ...
\end{lstlisting}
\caption{\emph{Refined decomposition in the search-and-rescue domain. ChatHTN must query for a decomposition of the task rescueSurvivor(John,Zulu). Our system
learns a method from the decomposition provided by ChatGPT for rescueSurvivor(Maria,Zulu)}, thereby avoiding a costly call to ChatGPT.}
\label{refinedHierarchy}
\end{figure}


\begin{table}[htb]
\centering
\renewcommand{\arraystretch}{1.2}
\begin{tabular}{|p{2.1cm}|p{3.7cm}|p{3.8cm}|p{3.8cm}|}
\hline
\textbf{Name} & \textbf{searchANDrescueM2} & \textbf{checkSurvivorsM2} & \textbf{rescueSurvivorM2} \\ \hline

\textbf{Arguments} &
\makecell[l]{?area} &
\makecell[l]{?loc} &
\makecell[l]{?survivor, ?loc} \\ \hline

\textbf{Preconditions} & 
\makecell[l]{area(?area),\\ location(?loc),\\ atLoc(?loc,?area),\\ weather(?loc,?clear),\\ not(scanned(?loc)),\\ not(safeZone(?loc))} & 
\makecell[l]{person(?survivor),\\ at(?survivor,?loc)} & 
\makecell[l]{isDrone(?drone),\\ safeHaven(?SH),\\ atDrone(?drone,?loc)} \\ \hline

\textbf{Subtasks} & 
\makecell[l]{!scanLocation(?loc),\\ checkSurvivors(?loc),\\ searchAndrescue(?area)} & 
\makecell[l]{rescueSurvivor(?survivor,\\ \ \ \ \ ?loc),\\ checkSurvivors(?loc)} & 
\makecell[l]{!pickUpSurvivor(?drone,\\ \ \ \ \ ?survivor,?loc),\\ !fly(?drone,?loc,\\
\ \ \ \ ?SH),\\ !unload(?drone,?survivor,\\
\ \ \ \ ?SH)} \\ \hline
\end{tabular}
\caption{Example HTN methods with their arguments, preconditions, and subtasks.
Question marks prefixes denote variables. }
\label{tab:htn-methods}
\end{table}

\section{HTN Planning}

An HTN planning problem is a 4-tuple \emph{(task list, state, methods, operators)}. Let $\tilde{t}=[t,t_1,\dots,t_n]$ be a task list and $s$ be the current state. HTN planning transforms $\tilde{t}$ and $s$ by recursively interleaving two steps, depending on whether $t$ is compound or primitive. During this process, a plan $\pi$ is generated, consisting of a list of primitive tasks. Initially, $\pi$ is an empty task list.

A method is a triple: \emph{(task, preconditions, subtasks)}, as exemplified in the three methods of Table \ref{tab:htn-methods}. Given a state $s$ (a collection of grounded atoms) and a compound task $t$, a method $m=(t_m,p_m,st_m)$ is applicable to $(s,t)$ if there is a variable substitution $\theta$ such that $t_m\theta = t$ and $p_m\theta$ holds in $s$. 
\footnote{A variable substitution $\theta$ is a mapping from variables to terms, such as $\theta=[(?x,1),(?y,a)]$. Applying $\theta$ to an atom $a$, written $a\theta$, substitutes any occurrence of the variable for the term. For instance, if $a=foo(?x,?y)$, then $a\theta = foo(1,a)$}
That is, if $p=\text{not}(p')$ is a negative precondition in $p_m$, then $p'\theta \notin s$,  
and if $p$ is a non-negative precondition in $p_m$, then $p\theta \in s$.

\paragraph{\textbf{If $t$ is compound}.} If a method $m=(t_m,p_m,st_m)$ is applicable to $(s,t)$ with a substitution $\theta$, then the task list $\tilde{t}$ is transformed into $\tilde{t} = st_m\theta \cdot [t_1,\dots,t_n]$.\footnote{The symbol $\cdot$ denotes the concatenation of two lists: $[t_1,\dots,t_n] \cdot [t'_1,\dots,t'_m]=[t_1,\dots,t_n,t'_1,\dots,t'_m]$.} In such a case, we say that $st_m\theta$ is a decomposition for $t$. The state $s$ remains unchanged when decomposing compound tasks. The current plan $\pi$ remains unchanged as well.

An operator is a 4-tuple \emph{(task, preconditions, add-list, delete-list)}, where the task is a primitive task, and preconditions have the same form as methods' preconditions, and the add-list and delete-list are lists of atoms. Given a state $s$ and a primitive task $t$, an operator $o=(t_o,p_o,add_o,del_o)$ is applicable to $(s,t)$ if there is a variable substitution $\theta$ such that $t_o\theta = t$ and $p_o\theta$ holds in $s$ (same definition as for methods).

\paragraph{\textbf{If $t$ is primitive}.} If an operator $o=(t_o,p_o,add_o,del_o)$ is applicable to $(s,t)$ with a substitution $\theta$, then the task list is transformed into $[t_1,\dots,t_n]$. The state $s$ is transformed into $o(s)$ as follows: $o(s) = (s \setminus del_o\theta) \cup add_o\theta$. This is referred to as applying $o$ to $s$. The plan $\pi$ is transformed into $\pi \cdot [t]$.
That is, the operator associated with $t$ is applied on $s$ and $t$ is appended to the end of the current plan.

The process continues until the task list $\tilde{t}$ is empty, in which case the current plan $\pi$ is returned. For instance, in Figure \ref{refinedHierarchy}, the plan consists of all primitive tasks in the order generated as shown in Figure \ref{fig:plan}.

\begin{figure}[htb]
\centering
\fbox{%
\begin{minipage}{0.92\linewidth}
\small\ttfamily
A plan for \emph{searchAndRescue(Alpha)}:\\
Primitive tasks:\ \emph{[!scanArea(Zulu), !fly(Drone01,safeHaven,Zulu),
            !pickUpSurvivor(Drone01,Maria,Zulu),
            !fly(Drone01,Zulu,safeHaven),
            !dropSurvivor(Drone01,Maria,SafeHeaven),
            !fly(Drone01,safeHaven, Zulu),
            !pickUpSurvivor(Drone01, John, Zulu),
            !fly(Drone01,Zulu, safeHaven),
            !dropSurvivor(Drone01,John, SafeHeaven),
            !doNothing(), $\dots$]}
\end{minipage}}
\caption{Plan for the hierarchy in Figure \ref{refinedHierarchy} decomposing \emph{searchAndRescue(Alpha)}.}
\label{fig:plan}
\end{figure}

\section{ChatHTN: Integrating ChatGPT and HTN Planning}

ChatHTN performs standard HTN planning as in the SHOP system \citep{nau1999shop}. However,
during the HTN planning process, if the current task list is  
$\tilde{t}=[t,t_1,\dots,t_n]$ and $s$ is the current state and no applicable method exists for $(s,t)$, then ChatHTN prompts ChatGPT for a decomposition by passing relevant information. However, in standard HTN planning, a compound task's semantics are defined by the methods that decompose it. This poses a problem since we need to specify to ChatGPT the precise meaning of what needs to be accomplished. To address this issue, we adopted the notion of annotated tasks from \citep{hogg2008htn}. An annotated task is a 3-tuple:  
\emph{(task, preconditions, effects)}. Figure \ref{fig:annotated-rescue} shows the annotated task for \emph{rescueSurvivor}.

\begin{figure}[htb]
\centering
\fbox{%
\begin{minipage}{0.92\linewidth}
\small\ttfamily
Annotated task:\\[2pt]
task:\ \emph{rescueSurvivor(?survivor,?loc)}\\
preconditions:\ \emph{safeHaven(?SH), at(?survivor,?loc)}\\
effects:\ \emph{(at(?survivor,?SH))}
\end{minipage}}
\caption{Annotated task specification for \emph{rescueSurvivor(?survivor,?loc)}.}
\label{fig:annotated-rescue}
\end{figure}
 
Annotated tasks give clear semantics to ChatGPT about what the task is trying to accomplish. In addition to the annotated task, ChatHTN also passes to ChatGPT the list of primitive tasks and their associated operators, as well as the current state $s$.

ChatGPT will, with some frequency, return an incorrect sequence of primitive tasks, $\tilde{t'}$, when decomposing $t$. There are two cases here. First, $\tilde{t'}$ cannot be executed on state $s$ because an associated operator in $\tilde{t'}$ is not applicable in the state. This is handled directly by the HTN planning process by backtracking. The second case is more problematic: $\tilde{t'}$ can be executed on state $s$, but the resulting state $s'$ does not satisfy the conditions on $t$'s effects. To address this issue, when decomposing a compound task $t$ in a task list $\tilde{t}=[t,t_1,\dots,t_n]$ with a decomposition $\tilde{t'}$ (i.e., generated either by using a method or by querying ChatGPT), ChatHTN adds a verifier task $t_{ver}$, so the resulting task list is: $\tilde{t'} \cdot [t_{ver}] \cdot [t_1,\dots,t_n]$.

A \emph{verifier task} $t_{ver}$ of a compound task $(t,p,add)$ is a primitive task whose associated operator has as preconditions $add$, and neither an add-list nor a delete-list. Its purpose is to check that the effects of $t$  are valid in the state after the HTN planning process removes $t$'s subtasks $\tilde{t'}$ and, recursively, all of its  subtasks  from the task list. As a result, ChatHTN is sound. Informally, the tasks in $\tilde{t}$ are satisfied by the plan generated by ChatHTN when called with $(s,\tilde{t})$. For a formal definition, please see \citep{munozavila2025chathtninterleavingapproximatellm}.


\section{Online Learning HTN methods}

As explained before, when ChatGPT returns a task list $\tilde{t'}$ decomposing a compound task $t$, with current task list $[t,t_1,\dots t_n]$ and current state $s$, the task list is modified as follows: $\tilde{t'} \cdot [t_{ver}] \cdot [t_1,\dots t_n]$.

To better explain the method learning process, let $\tilde{t'} = [t'_1,\dots,t'_m]$ (and hence, the task list is $[t'_1,\dots,t'_m, t_{ver}, t_1,\dots,t_n]$). Let $o'_i$ be the associated operator for primitive task $t'_i$. Let $[s_0,s_1,\dots,s_m]$ be the sequence of states generated, i.e.,  
$s_0=s$, $s_1=o'_1(s_0)$, and so forth until $s_m = o'_m(s_{m-1})$ and  
$s_m=s'$.  
The new state $s'$ generated after applying in this way $[o'_1,\dots,o'_m]$ to $s$ is denoted as $apply([o'_1,\dots,o'_m],s)=s'$.

Assuming all the operators $o'_i$ are applicable in their corresponding states $s_{i-1}$, ChatHTN then checks $t_{ver}$ on $s'$. That is, it checks if the preconditions of $t_{ver}$ are valid in $s'$. In doing so, it checks that the effects, $eff_t$, of the annotated task for $t$ are valid in $s'$.

Online learning of the method occurs immediately after this verification succeeds: a new method is learned $(t{\uparrow},p{\uparrow},\tilde{t'}{\uparrow})$. Given an atom $a$ with no variables, such as $at(Maria,Zulu)$, a lifted atom, denoted by $a{\uparrow}$, is the atom with all constants replaced by variables. For instance, $at(Maria,Zulu){\uparrow}$ $= at(?Maria,?Zulu)$. Thus, both $t$ and all atoms in $t$'s decomposition, $\tilde{t'}$, are lifted in the learned method.

Finally, $p$ is the set of preconditions generated by performing goal regression \citep{mitchell1986explanation} from $eff_t$ on the sequence $(o'_1,\dots,o'_m)$, written $Reg([o'_1,\dots,o'_m],eff_t)$, which computes the minimum conditions needed in any state $s"$ for the sequence to be applicable. This is precisely what we need in order to safely lift the variables and apply the task decomposition in other states: to ensure that when the preconditions of the learned method are applicable in any state $s"$, after ChatHTN processes the subtasks $[t'_1,\dots,t'_m]$, the effects of $t$ are valid in the resulting state, 
$apply([o'_1,\dots,o'_m],s")$. Regression, $Reg([o'_1,\dots,o'_m],eff_t)$, is defined as follows:

\[
Reg([o'_1,\dots,o'_i],g) =
\begin{cases}
(g_i - add_i) \cup pre_i, &
\parbox[t]{0.55\linewidth}{\raggedright
if $i \ge 1$ and $i < m$, and $add_i$, $pre_i$ are the add-list\\
and preconditions when applying $o'_i(s_{i-1})$,\\
and $g_i = Reg([o'_{i+1},\dots,o'_m],g)$}\\[4pt]
eff_t, &
\parbox[t]{0.55\linewidth}{\raggedright
if $i = m$}
\end{cases}
\]

Once $p = Reg([o'_1,\dots,o'_m],eff_t)$ is generated, we lift the atoms in $p$.\footnote{We are assuming that the preconditions and effects of the operators are all lifted. If some are not lifted, the regression definition must be changed to account the non-lifted arguments in preconditions or effects, so these arguments are also not lifted in $p{\uparrow}$.}

In addition to the methods learned as described, we also provide the \emph{termination methods.} Given an annotated task, \emph{(t, p, effects)}, the termination method, $m_t$, for $t$ is defined as $m_t=(t,\text{effects},(!doNothing()))$. That is, the method has as preconditions the annotated task's effects and a single subtask: the primitive task \emph{doNothing}. The associated operator for this subtask has no preconditions (so it is always applicable) and has no effects, so it doesn't change the state. Its use is to stop recursive calls for $t$ from other methods, in states where the desired effects of the task are satisfied. For instance, in Figure \ref{refinedHierarchy}, the last \emph{checkSurvivors(Zulu)} task is decomposed into  
\emph{!doNothing()}, because there are no more survivors in Zulu.  
Figure \ref{fig:terminationMethod} shows the termination method for \emph{checkSurvivors(?loc)}.

\begin{figure}[htb]
\centering
\fbox{%
\begin{minipage}{0.92\linewidth}
\small\ttfamily
Termination method:\\[2pt]
task:\ \emph{checkSurvivors(?loc)}\\
preconditions:\ \emph{not(at(?survivor,?loc))}\\
subtask:\ \emph{doNothing()}
\end{minipage}}
\caption{termination method for \emph{checkSurvivors(?loc)}.}
\label{fig:terminationMethod}
\end{figure}


\section{Experiments}

In our  experiments we are using GPT-4-turbo, We selected GPT-4-turbo for its demonstrated performance on symbolic and multi-step reasoning tasks, and for its cost-efficiency, which enabled extensive experimentation. The model’s 128K context window was particularly useful for representing hierarchical task networks and long planning traces.

\subsection{Domains}

\paragraph{Logistics Transportation Domain} The Logistics Transportation domain is a  benchmark in HTN planning that simulates the process of transporting packages using different vehicles \citep{veloso1993derivational}. Trucks can move between different locations within the same city, while airplanes can travel between different cities, but only between airports. The goal is to transfer a package from source location to destination location.

\paragraph{Search and Rescue Domain}The Search and Rescue domain is a benchmark in HTN planning that simulates the process of locating and rescuing victims in a hazardous environment. A drone can fly to different locations, where it can scan the area and rescue one survivor at a time. The goal is to survey all locations and rescue all survivors.

\subsection{Performance Measure}
We use the following two metrics to measure the performance and compare between the planner with and without method learner: (1) the number of calls to GPT; (2) the percentage of problems solved.

\subsection{Experimental Settings}

\paragraph{Common settings} We handcrafted the HTN knowledge base. This knowledge base consists of operators, methods, and annotated tasks such that a correct solution can be generated for any given solvable problem. To create a testbed, we randomly generated 10 problems. To simulate an incomplete domain, we removed one method at a time and attempted to solve the 10 problems. Thus, some of the problems might not be solvable because of the absence of the method. After each instance — in which a problem is given, regardless of whether ChatHTN succeeds in finding a solution or not, the methods learned during that instance are deleted. For each problem and each method removed, we run the problem 3 times and compute the average of the two performance measures. We repeated this for every method in the domain. Therefore, we tested our learning procedure both when methods are missing for low-level tasks in the hierarchy such as \emph{rescueSurvivor}, as well as methods for high-level tasks such as \emph{searchAndRescue}. The latter may require ChatGPT to generate a solution for the whole problem. We tested our HTN learning process in two domains, which we describe next.

\paragraph{Logistics Transportation Domain.}
The logistics transportation domain's main task is to relocate a package. If the package is to be relocated within the same city as its starting location, a truck can be used to relocate the packages. However, if the pacdkage needs to be relocated between locations in different cities, a combination of truck transport and airplane transport is needed. Trucks can only transport packages between locations in the same city whereas airplanes can transport packages between cities but they only fly from and to airports. 

The following is a short description of the methods in this domain. {\bf  TM1} is the termination method for truck transportation. {\bf  TM2} transfers a package from a source location to a target location if a truck and the package are in the same location. {\bf  TM3} transfers a package from a source location to a target location when a truck and the package are in the different locations. {\bf  AM1} is the termination method for airplane transportation. {\bf  AM2} transfers a package from a source location to a target location if an airplane and the package are in the same airport. {\bf  AM3} transfers a package from a source location to a target location when an airplane and the package are in the different locations. {\bf TPM1} transfers a package within one city. {\bf TMP2} transfers a package between two cities.

The Logistics Transportation domain is structured around a network of three cities, each containing two post offices and an airport. There are three trucks, one in each city, and one airplane. A problem instance is defined by a task list of 5 tasks, where each task involves moving a package from a source location to a destination location. When a planning succeeds, plans have a variety of lengths with as few as in the  low 50s and as many as in the high 70s.

\paragraph{Search and Rescue Domain.} 
In the Search and Rescue domain, the environment consists of a set of distinct locations categorized as either unsafe zones or a single safe zone serving as the rescue destination. A drone is capable of flying between any two locations, scanning an area to detect survivors, and transporting one survivor at a time; however, it must scan a location before performing any subsequent actions like checking for or rescuing them. The initial state contains one drone and five survivors randomly distributed among the unsafe zones. A problem consists of single \emph{searchAndRescue} tasks, which requires rescuing all five survivors. When a planning succeeds, plans have a variety of lengths with as few as in the  low 21 and as many as in the high 47.  

The following is a short description of the methods in this domain. {\bf  SCAN1} is the termination method for scanning one location. {\bf  SCAN2} scans a location if a drone is at that location. {\bf  SCAN3} scans a location when a drone is at a different location. {\bf CS1} is the termination method for checking survivors at one location. {\bf CS2} checks survivors at one location and rescues them. {\bf RS1} rescues a survivor at a location if a drone is at the same location. {\bf RS2} rescues a survivor at a location when a drone is at a different location. {\bf SAR1} is the termination method for search and rescue. {\bf SAR2} searches and rescues all the survivors.
\subsection{Results}

\begin{figure}[htb]
    \centering
        \includegraphics[width=\textwidth]{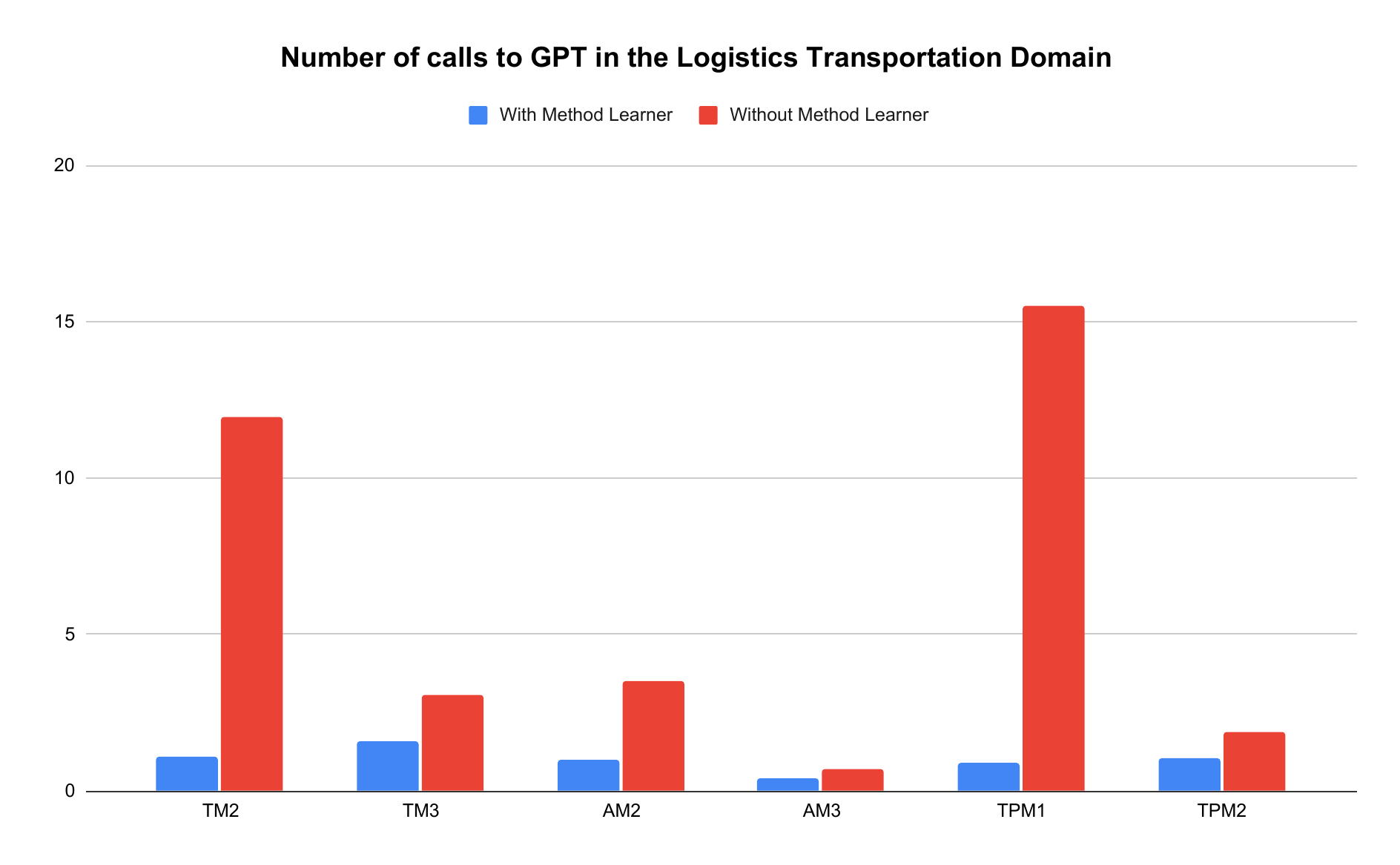}
        \caption{This plot compares number of calls to ChatGPT of the planning system with and without method learner in the Logistics Transportation Domain). The y-axis is the number of calls to ChatGPT; the x-axis indicates the names of the methods removed for the method learner. The TM methods are methods for truck transportation; The AM are methods for air transport; and the TPM methods are the high-level transport package methods.}
        \label{fig:callsLogistics}
 \end{figure}

 \begin{figure}[htb]
    \centering
    \includegraphics[width=\textwidth]{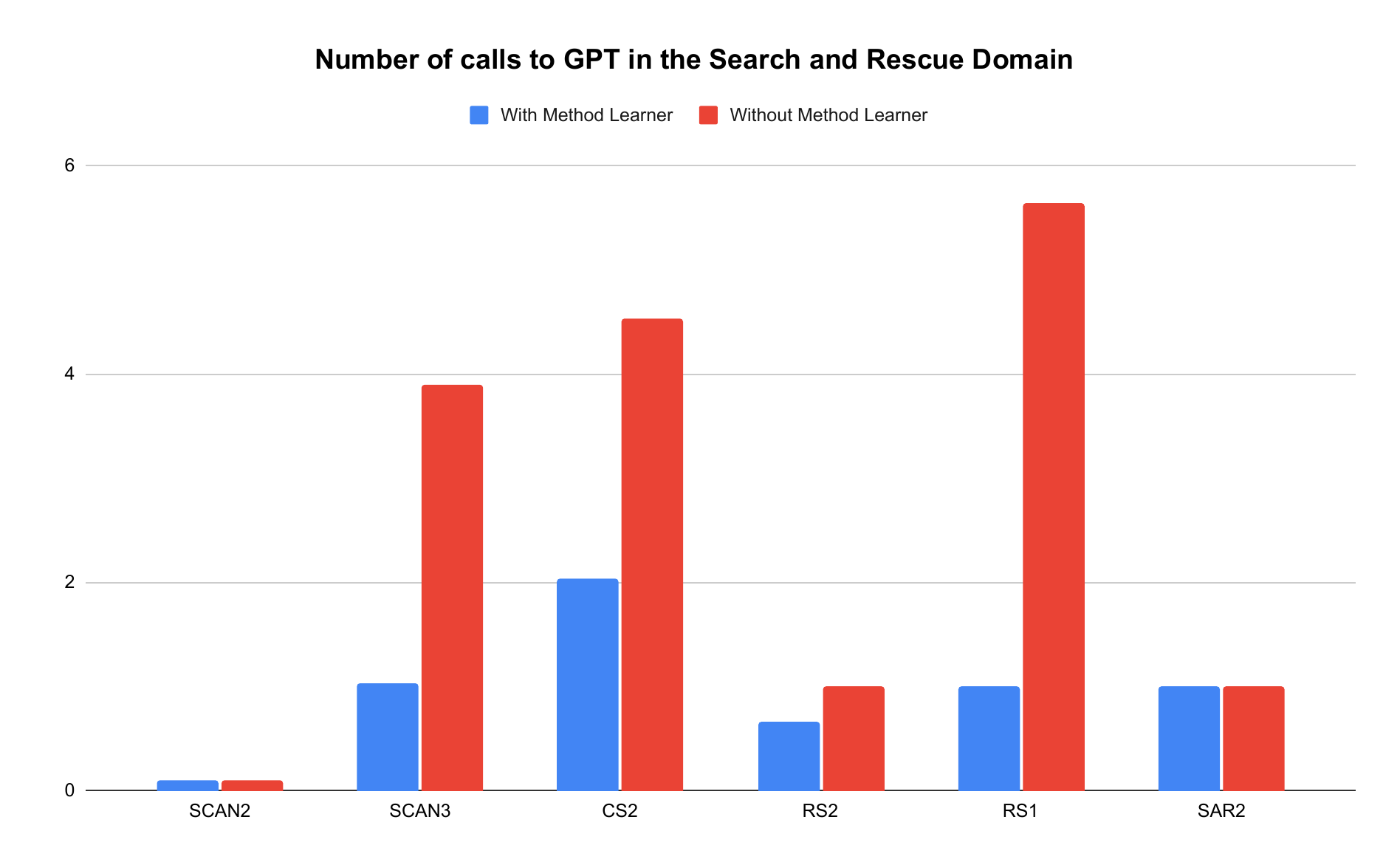}
    \caption{This plot compares number of calls to ChatGPT of the planning system with and without method learner in the Search and Rescue Domain). The y-axis is the number of calls to ChatGPT; the x-axis indicates the names of the methods removed for the method learner. The scan methods are for the scan location task; the CS methods for the checkSurvivors task; the RS is for the rescueSurvivor task; and SAR is for the high-level searchANDrescue task.}
    \label{fig:callsSearch}
\end{figure}

\begin{figure}[htb]
    \centering
        \includegraphics[width=\textwidth]{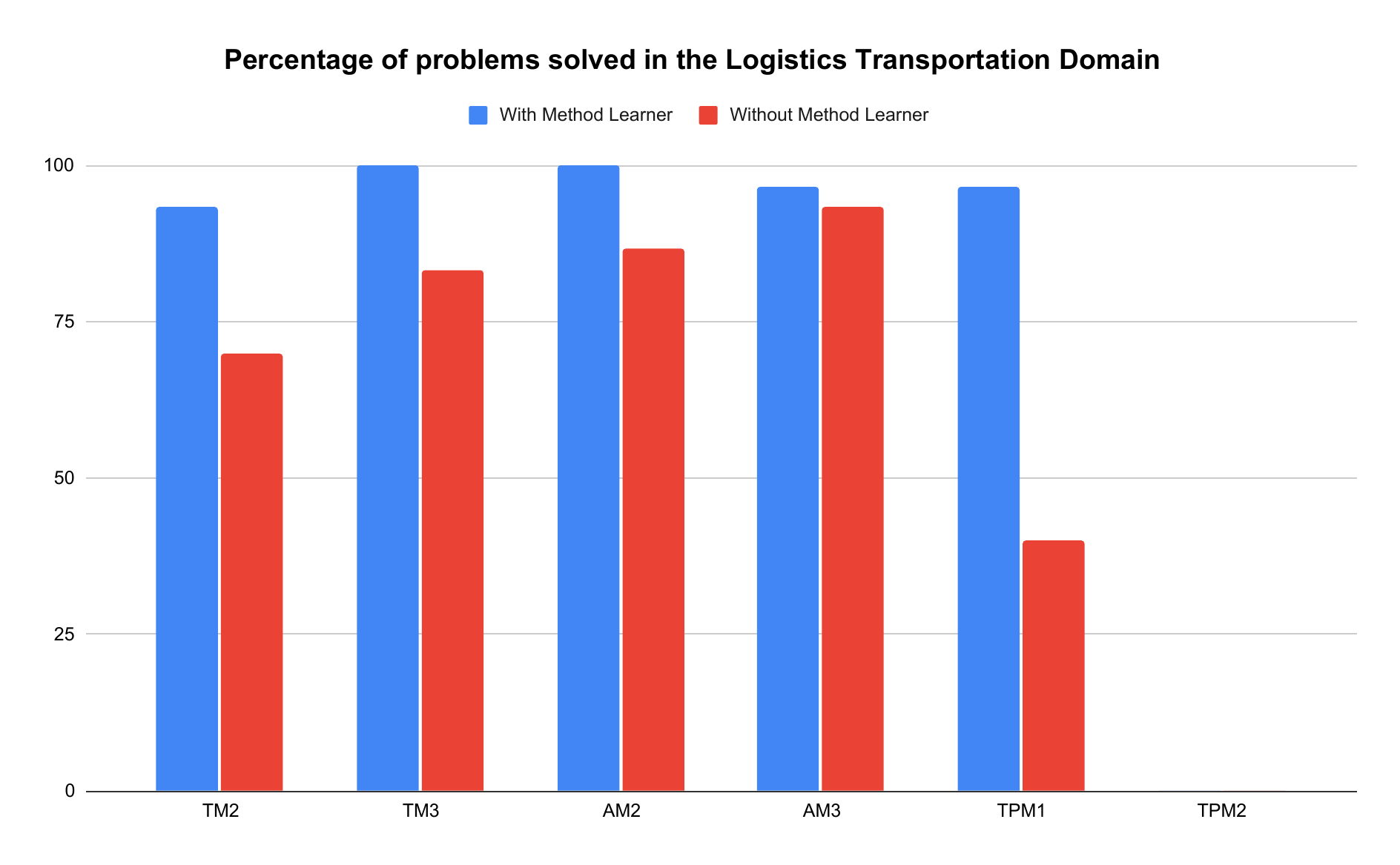}
        \caption{This plot compares the percentage of problems solved by the planning system with and without method learner in the Logistics Transportation Domain). The y-axis is the percentage of problems solved; the x-axis indicates the names of the methods removed for the method learner. The TM methods are methods for truck transportation; The AM are methods for air transport; and the TPM methods are the high-level transport package methods.}
        \label{fig:solvedLogistics}
    \end{figure}

\begin{figure}[htb]
    \centering
        \centering
        \includegraphics[width=\textwidth]{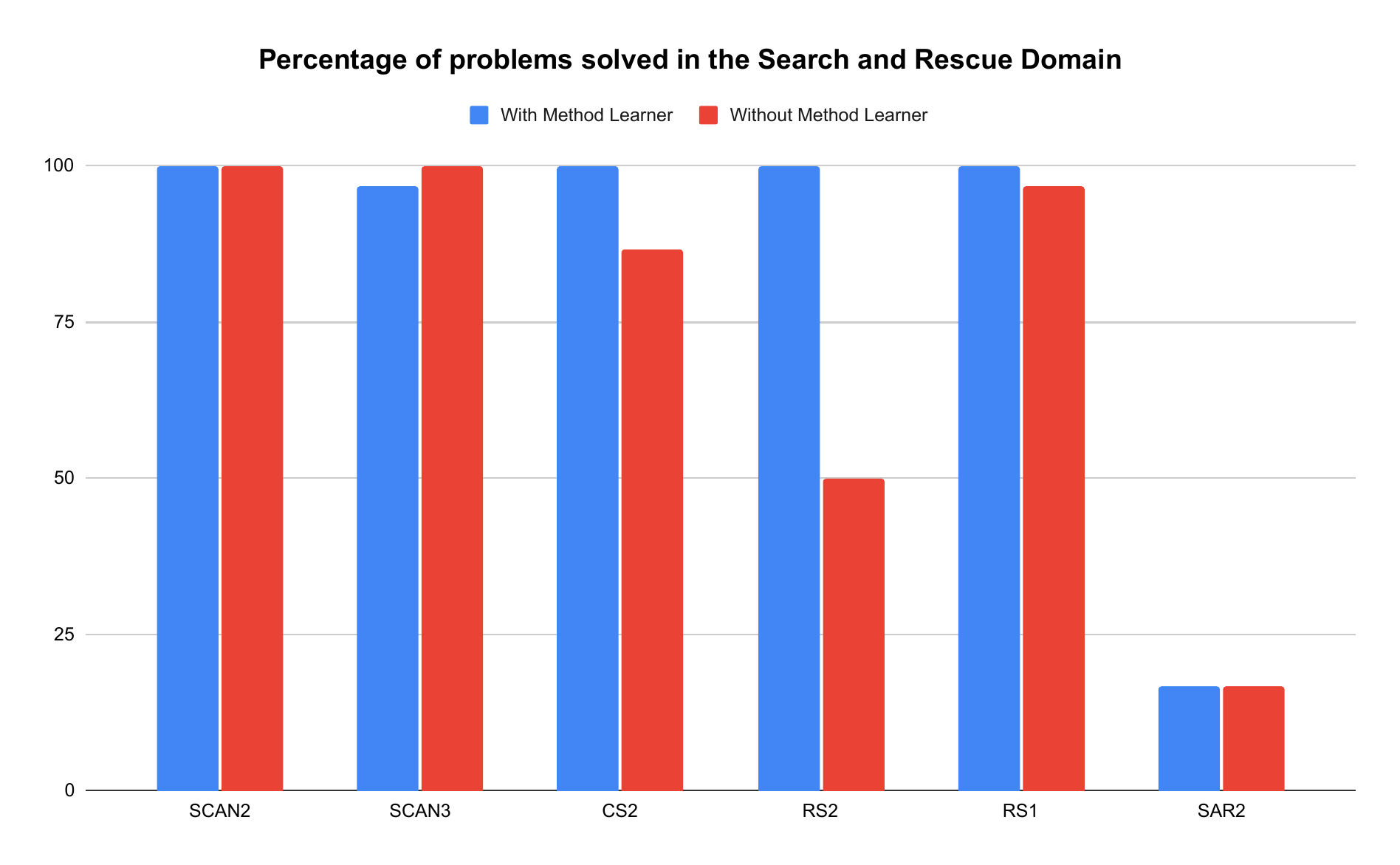}
     \caption{This plot compares the percentage of problems solved by the planning system with and without method learner in the Search and Rescue Domain). The y-axis is the percentage of problems solved; the x-axis indicates the names of the methods removed for the method learner. The scan methods are for the scan location task; the CS methods for the checkSurvivors task; the RS is for the rescueSurvivor task; and SAR is for the high level searchANDrescue task.}
    \label{fig:solvedSearch}
\end{figure}

Figure \ref{fig:callsLogistics} compares of the average number of calls made to ChatGPT to solve the 10 randomly generated problems for the Logistics Transportation and Figure \ref{fig:callsSearch} shows the results for the Search and Rescue domain. The red and blue bars indicate the performance of the planner without and with the method learner, respectively.

The main observation from these results is that employing the method learner consistently leads to a reduction in the number of calls to ChatGPT. This improvement stems from the learner's ability to generalize from each interaction with ChatGPT. When the planner encounters a task it cannot decompose, it queries ChatGPT. Our method learner then processes the specific action sequence returned by the LLM and formulates a generic, reusable HTN method. This learned method is applicable to the same task decomposition that may arise later in the planning process, thereby averting the need for redundant queries.

In contrast, the planner without the method learner is unable to leverage "past experience". When faced with a task for which ChatGPT has previously provided a decomposition, the planner cannot generalize that specific solution into a generic method. Consequently, it must re-query ChatGPT for every new instance of the task, leading to inefficiency. Note that the method learner is different from a memoization module: caching a sequence of actions for a given state is insufficient, as this knowledge fails to apply to different states under the same decomposition. The key advantage of our method learner is its ability to synthesize abstract, reusable methods from these concrete, state-specific examples provided by the LLM.

Figure \ref{fig:solvedLogistics} compares the percentage of problems solved across the 10 randomly generated problems for the Logistics Transportation Domain and Figure \ref{fig:solvedSearch} shows the results for the  Search and Rescue domains. The red and blue bars indicate the performance of the planner without and with the method learner, respectively.

We can make two further observations. First, neither configuration guarantees a 100$\%$ success rate in all scenarios. This is expected, as ChatGPT does not ensure the correctness of its generated action sequences, and therefore the verifier tasks return a failure. Our system's verification step is crucial here, as it validates the ChatGPT's output, ensuring that the planner does not proceed with an incorrect plan and that the method learner does not acquire a faulty method from an erroneous response. Second, the success rate is frequently lower for the planner without the method learner. This is because each time the planner queries the ChatGPT for a task decomposition, there is a possibility of receiving an incorrect response. By repeatedly querying for similar tasks instead of learning from the first successful interaction, the planner increases its exposure to potential LLM errors, thus increasing the cumulative probability of failure.

The empirical results also highlight a significant challenge for the planner when high-level methods are missing from the knowledge base. This is particularly evident when observing the performance after removing the highest-level methods in each domain—TPM2 in the Logistics domain and SAR2 in the Search and Rescue domain (Figures \ref{fig:solvedLogistics} and \ref{fig:solvedSearch}). The reason for this low success rate is that the planner, lacking an initial method to decompose the top-level task, must immediately query ChatGPT to generate a solution for the entire problem. This effectively tasks ChatGPT with generating a complete sequence of primitive actions from the initial state. Requiring an LLM to produce a long, entirely correct action sequence is a substantially more complex task, which is known to be prone to errors.

A key characteristic of the current learning procedure is that it generates methods composed exclusively of a linear sequence of primitive tasks. This design choice means the system cannot learn more general methods that include compound subtasks, which are necessary for implementing abstract control structures like recursion or iteration. This limitation can be illustrated with an example. The $\emph{checkSurvivors(loc)}$ method shown in Table 1 uses recursion to check and rescue an arbitrary number of survivors at a location. Our current learning procedure cannot generate such a method. Instead, if it learns from a ChatGPT decomposition for a location containing two survivors, it will generate a specialized method for rescuing exactly two survivors. This learned method is brittle; it cannot be applied to a new situation involving a larger number of survivors. Consequently, the planner must re-query ChatGPT to learn a new, distinct method for each unique number of survivors it encounters, rather than generalizing a single, recursive solution.

\section{Related Work}

The topic of learning HTNs is a recurrent theme in the research literature. More recently, \citep{langley2025learning} presents an overview that examines differing assumptions systems make about the types of knowledge provided. Specifically, it points to forms of inputs for the learner, in our case being the decomposition of a task into a list of primitive tasks. It also points to a variety of learning objectives, ours being generalizing the given decomposition as well as learning applicability conditions. We select a few exemplary works; for a more comprehensive review, see \cite{langley2025learning}.

\cite{reddy1997learning} is one of the earlier HTN learners. Although not using an HTN planner, it uses so-called d-rules to decompose goals into subgoals (akin to hierarchical goal networks, HGNs \citep{shivashankar2012hierarchical}, which represent goal-subgoal decompositions instead of task-subtasks decompositions in HTNs). It uses inductive generalization to learn goal decomposition constructs, which relate goals, subgoals, and conditions for applying these d-rules. By grouping goals in this way, hierarchical models are learned that lead to speed-ups in problem-solving. However, it is possible to solve the same problems without the learned task models. In our work, we are interested in learning methods for general HTN planning, which is strictly more expressive than STRIPS planning \citep{erol1994htn}.

Our work is related to learning HTNs through teleoreactive logic programs \citep{choi2005learning}. The key idea is that the HTN planner performs a hierarchical decomposition and encounters a gap in its HTN knowledge between a state $s$ generated so far and some desired state $s'$. In such a situation, it calls a first-principles planner to generate a plan from $s$ to $s'$, and learns a hierarchical structure represented as a collection of Horn clauses describing the tasks. These Horn clauses serve a similar role to the annotated tasks, providing semantics for the tasks. In our work, the decompositions are provided by ChatGPT, and we additionally provide explicit soundness guarantees for the resulting system, which is needed because, unlike a first-principles planner, LLM-based chatbots and large-reasoning models can provide incorrect solutions \citep{shojaee2025illusion} and have no means of self-verifying the correctness of their results \citep{stechly2024self}.

Annotated tasks were introduced in \cite{hogg2008htn} for the purpose of learning hierarchical decompositions. However, the learning problem differs: in that work, the system is provided with state–action plan traces and learns a set of methods from these traces and annotated tasks. In contrast, our system is incremental, learning methods as they are needed. That work was extended in a number of settings for HTN learning, including nondeterministic domains \citep{hogg2009learning} and domains where actions have associated costs \citep{hogg2010learning}.

A related area of research investigates interactive task learning (ITL) systems that learn symbolic knowledge structures through natural language dialogue between humans and the agent. The ROSIE system learns symbolic task knowledge by asking a human teacher for definitions of unknown concepts \citep{kirk2019learning}. The VAL system uses an LLM as a parser to interpret user instructions and recursively clarifies unknown actions to build an HTN structure \citep{lawley2024val}, while the STARS framework uses an agent to analyze and repair multiple LLM-generated plans before seeking simple confirmation from a human \citep{kirk2024improving}. 
The main difference is that our work does not rely on natural language dialogue with a human for task decomposition and verification. Instead, our system queries LLM for a task decomposition and then learns a generalized method from it; by using goal regression to derive the method’s preconditions, we guarantee it is correct and reusable, thereby reducing the need for future LLM queries.

Other works aim at learning both the operators and methods simultaneously in domains that are partially observable \citep{zhuo2014learning}. As in the previous system, the learning problem is defined by being given a collection of state–action plan traces and annotated tasks. The difference from the previous system is that the intermediate states are only partially observed (i.e., there are missing atoms in the state). Interestingly, this work shows that learning methods and operators simultaneously is more effective than first learning the methods and then the operators. Again, the main difference from our work is that our learner is incremental, learning methods as needed, although we assume full state observability.

Other works also aim at learning the goals for HGNs and the HTN methods \citep{fine2020learning}. That system receives as input the operators and traces but not the task semantics. It casts the input traces as sentences (e.g., each atom as a word) and uses word embeddings to group similar atoms, treating them as goals in HGNs. It also extracts structure (i.e., the goal–subgoal decompositions). Furthermore, it learns numerical conditions. 

Researchers have also investigated using LLMs for authoring planning models e.g., \citep{oates2024llmplanning}, including hierarchical planning models  \citep{fine-morris2025leveraging}. 
In the latter work, the authors propose a pipeline that process documents to generate the methods. The pipeline progressively elicits a hierarchical structure, starting with the documents and some background knowledge. In the intermediate stage it generates semistructured natural language task decompositions. The final result is a collection of methods  for Hierarchical Goal Networks (HGNs) \citep{oates2024llmplanning}. It is conceivable that such work could be combined with ours, in that that work is used as the offline learner of the initial collection of methods, and ours is used to refine the methods online while applying them during planning.

\section{Conclusion}

In this paper, we presented a procedure for online learning of HTN methods. It is integrated with the ChatHTN planner. ChatHTN performs standard HTN planning, decomposing the first task $t$ in its current task list depending on the current state. However, unlike standard HTN planning,  if $t$ is compound and no method exists decomposing $t$, ChaTHTN queries ChatGPT for a decomposition of $t$ into a sequence of primitive tasks $\tilde{t}$, followed by a verifier task, $t_{ver}$, that checks if $t$'s effects are satisfied after processing $\tilde{t}$. In such a situation, goal regression is performed to obtain a method decomposing the generalized task $t{\uparrow}$ into the generalized task sequence $\tilde{t}{\uparrow}$. Goal regression guarantees that whenever the method's preconditions are valid in an state $s$ and a task matching $t{\uparrow}$, then after ChatHTN processes the instance of $\tilde{t}{\uparrow}$, it will result in an state satisfying $t_{ver}$.
Termination methods, which decompose $t$ into the dummy task \emph{doNothing}, check if $s$ already satisfies the effects of  $t_{ver}$, allowing ChatHTN to continue planning for other tasks in the task list.

We tested ChatGPT with the online method learner against standard ChatGPT in the Logistics Transportation  and in the Search and Rescue domains. In both domains we observed that with online method learning reduces the number of calls to ChatGPT while solving at least as many problems, and in some cases, even more.

We will explore the following directions for future work. Currently, ChatHTN queries ChatGPT to decompose the current task into a sequence of primitive tasks. We want to modify ChatGPT to allow the current task to be decomposed into a sequence of compound and primitive tasks. In doing so, we expect to address a limitation of our method learner, which cannot learn recursive methods. Another alternative way to learn recursive methods is to analyze the sequence of primitive tasks $\tilde{t}$ returned by ChatHTN and use a mechanism similar to \cite{hogg2008htn} taking advantage of the annotated tasks to induce a hierarchical structure (i.e., when $\tilde{t}$ repeats patterns of subsequences such as \emph{load,drive,unload,load,drive,unload}.

 


{\parindent -10pt\leftskip 10pt\noindent
\bibliographystyle{cogsysapa}
\bibliography{format}

}


\end{document}